\documentclass[sigconf]{acmart}
\usepackage{booktabs} 
\usepackage{amssymb,amsmath,subcaption}
\usepackage{balance}
\usepackage{algorithm}
\usepackage{algorithmic}
\usepackage{epstopdf}
\newcommand{\remove}[1]{}

\usepackage{amssymb}
\usepackage{graphicx}
\usepackage{color}
\newcommand{\xhdr}[1]{\vspace{1mm}\noindent{{\bf #1.}}} 
\newcommand{\dnote}[1]{\textcolor{blue}{$\ll$\textsf{#1 --Dafna}$\gg$}}

\usepackage{hyperref}
\usepackage{array}
\newcolumntype{L}{>{\centering\arraybackslash}m{3cm}}
\usepackage{wrapfig}
\usepackage{paralist}

\setcopyright{acmcopyright}

\begin{document}

\title{Ballpark Crowdsourcing: The Wisdom of Rough Group Comparisons}


\author{Tom Hope}
\affiliation{%
  \institution{The Hebrew University of Jerusalem}
}
\email{tom.hope@mail.huji.ac.il}

\author{Dafna Shahaf}
\affiliation{%
  \institution{The Hebrew University of Jerusalem}
}
\email{dshahaf@cs.huji.ac.il}

\begin{abstract}

Crowdsourcing has become a popular method for collecting labeled training data. However, in many practical scenarios traditional labeling can be difficult for crowdworkers (for example, if the data is high-dimensional or unintuitive, or the labels are continuous).  

In this work, we develop a novel model for crowdsourcing 
that can complement standard practices by exploiting people's intuitions about \emph{groups} and \emph{relations} between them. 
We employ a recent machine learning setting, called \emph{Ballpark Learning},
that can estimate individual labels given only coarse, aggregated signal over groups of data points. 
To address the important case of continuous labels, we extend the Ballpark setting (which focused on classification) to regression problems. We formulate the problem as a convex optimization problem and propose fast, simple methods with an innate robustness to outliers.

We evaluate our methods on real-world datasets, demonstrating how useful constraints about groups can be harnessed from a crowd of non-experts. Our methods can rival supervised models trained on many \emph{true} labels, and can obtain considerably better results from the crowd than a standard label-collection process (for a lower price). By collecting rough guesses on groups of instances and using machine learning to infer the individual labels, our lightweight framework is able to address core crowdsourcing challenges and train machine learning models in a cost-effective way.

\end{abstract}

%
%


\remove{
\acmPrice{15.00}
\acmDOI{10.1145/3159652.3159670}
\acmISBN{978-1-4503-5581-0/18/02}

\acmConference[WSDM 2018]{WSDM 2018: The Eleventh ACM International
Conference on Web Search and Data Mining}{February 5--9, 2018}{Marina
Del Rey, CA, USA}
\acmYear{2018}
\copyrightyear{2018}
}
\remove{
\copyrightyear{2018} 
\acmYear{2018} 
\setcopyright{acmcopyright}
\acmConference[WSDM 2018]{WSDM 2018: The Eleventh ACM International Conference on Web Search and Data Mining }{February 5--9, 2018}{Marina Del Rey, CA, USA}
\acmPrice{15.00}
\acmDOI{10.1145/3159652.3159670}
\acmISBN{978-1-4503-5581-0/18/02}
}

\copyrightyear{2018} 
\acmYear{2018} 
\setcopyright{acmcopyright}
\acmConference[WSDM 2018]{WSDM 2018: The Eleventh ACM International Conference on Web Search and Data Mining }{February 5--9, 2018}{Marina Del Rey, CA, USA}
\acmBooktitle{WSDM 2018: WSDM 2018: The Eleventh ACM International Conference on Web Search and Data Mining , February 5--9, 2018, Marina Del Rey, CA, USA}
\acmPrice{15.00}
\acmDOI{10.1145/3159652.3159670}
\acmISBN{978-1-4503-5581-0/18/02}

\maketitle

\section{Introduction}


In many real-world learning scenarios, acquiring labeled training data is a challenging bottleneck for researchers and practitioners. 
%
Crowdsourcing has become a popular approach for
annotating large quantities of data. Platforms such as Amazon's Mechanical Turk allow researchers to distribute labeling tasks to a large number of crowdworkers, resulting in an effective mechanism for annotating data for supervised
learning models \cite{vondrick2010efficiently,rashtchian2010collecting}.

Despite its many benefits, crowdsourcing is not a panacea for data labeling. Crowdworkers are not domain \emph{experts}, and often are careless or make mistakes, generating unreliable labels. Since crowdworkers are error-prone, it is common to ask multiple workers to label each data point, which might make the labeling process prohibitively expensive.  In particular, crowds tend to be \emph{biased} \cite{whalen2015using}, requiring special methods to de-bias their answers. 

In addition, some tasks cannot be crowdsourced due to inherent \emph{hardness} of the task. The input might not be interpretable to most people (e.g., medical EEG data), or the output might be difficult to assess (e.g., continuous targets are notoriously hard for annotators \cite{whalen2015using}). Other times one cannot show people individual data points for labeling, especially due to \emph{privacy} considerations.




In this paper we propose a different approach to crowdsourcing, which could mitigate these problems. 
Rather than collecting labels for individual data points, we focus on collecting {\bf coarse, aggregated information} over \textbf{groups of points}, and using this information to {\bf infer individual labels}. Instead of investing resources in trying to get precise labels, we propose a lightweight framework that pools noisy crowd guesses over group averages and comparisons, and uses recent advances in machine learning to turn them into instance-level predictions.



For example, consider an advertising company. To optimize ad display, demographics are widely used to characterize customers. However, in practice such information (e.g., age, gender or richer targets like medical conditions) is usually unavailable \cite{dong2014inferring}.

Despite not having this information, the company might have access to millions of user-behavior mobile traces: session data, geo-location information, data derived from motion sensors, device specs, connection data, and more. These patterns are hard to interpret, rendering the data very difficult to label. Even if the company could solve the glaring privacy issues, it is left with potentially many millions of high-dimensional, complex pieces of information about users' mobile usage. Labeling this kind of data instance-by-instance is likely to be a painstaking, error-prone process \cite{zhao2011robust}. 



Instead of trying to label individual users, it might be easier to obtain some coarse signals on \emph{groups} of users. For example, Millennials text more and talk less \cite{forbes2015millenials}. People who tend to stay up and wake up late are more likely to be single \cite{maestripieri2014night}. 
Advertisers could, of course, be interested in going beyond  basic demographics, wishing to learn about health issues or political leanings, which are not directly reflected in mobile usage. This renders the attempt to label individual mobile usage patterns even more problematic \cite{zhao2011robust}, while getting estimates on groups may be far more feasible.



We take advantage of a new machine learning setting we have proposed in \cite{hope2016ballpark}, called \textbf{Ballpark Learning}. 
In the Ballpark setting we have unlabeled instances divided into bags, and we are given some aggregate information about label averages in bags in the form of loose constraints. For example, the bag of people who barely text has a higher average age than the bag of people who text often. Using only this kind of aggregate information, the goal is to predict \emph{individual labels} -- i.e., demographics of individual users. 


We suggest that the Ballpark setting is particularly useful for crowdsourcing: 
Instead of asking the crowd to label particular users, we could construct bags of users based on some simple attributes (e.g., monthly volume of text messages) and ask people to guess which bag has older users, and by how much. 


Using Ballpark learning can help address the shortcomings of crowdsourcing discussed above. Coarse guesses on simple groups require less expertise than individual labels, and fewer questions asked, as each question provides information for many datapoints.  
The datapoints we show the crowd are interpretable, since we can focus on a few intuitive dimensions and let the machine learning method take advantage of the other, less intuitive ones. The need to de-bias the crowd is also less pronounced, because our approach only requires wide intervals around the true average.
Finally, there are no privacy concerns, as we never show crowdworkers individual datapoints. (Interestingly, estimating individual labels from group statistics does have important implications regarding privacy, as Ballpark techniques can be used on sensitive data; however, crowdworkers are only exposed to aggregated data.)


Our key contributions are:
%
%
%

{ \setdefaultleftmargin{2em}{3em}{}{}{}{}
\begin{compactitem}
\item We propose a new model for crowdsourcing that can complement
standard practices by exploiting people's intuitions
about groups and relations between them. We exploit the natural
human tendency for intuiting on groups and the tendency
for comparison \cite{thurstone1927law} to glean interesting, informative patterns.
To the best of our knowledge, we are the first to focus on
comparisons between groups as an important part of labeling.
%
\item We adapt our recent machine learning setting, Ballpark Learning,
to crowdsourcing, and demonstrate its effectiveness. To
address the important case of continuous labels (which are
notoriously hard for crowdworkers), we extend the Ballpark
setting into \emph{regression}. We formulate the Ballpark regression
problem as a convex optimization problem and present fast,
simple methods to solve it with a natural robustness to outliers
that compares favorably to robust regression techniques.
%
    \item We demonstrate our results on real-world datasets and show
that by using weak constraints harnessed from a crowd of
non-experts, our methods are able to achieve results that rival
supervised models based on many true labels. We discuss
various ways to query the crowd for these constraints. In our
experiments, we obtained better results than those reached by
standard label collection -- at less than a third of the price.
\end{compactitem}
}

\section{Problem Formulation}
\label{sec:Formulation}
We now present the Ballpark Learning framework. In the following sections, we will demonstrate its usefulness for crowdsourcing. 

Consider a set of $N$ training instances $\mathcal{X}_N = \{ \mathbf{x}_1, \mathbf{x}_2,\ldots, \mathbf{x}_N \}$.  Each $\mathbf{x}_i$ has a corresponding \textit{unknown} label  $y^{*}_i \in \mathcal{Y}$. We extend our previous work \cite{hope2016ballpark} and allow label space $ \mathcal{Y}$ to be discrete or continuous, depending on the setting.
Along with unlabeled instances $\mathcal{X}_N $, we could be given a (possibly empty) set of $L$ labeled training instances  $\mathcal{X}_L = \{ \mathbf{x}_{N+1}, \ldots, \mathbf{x}_{N+L} \}$ with known targets $y_i$, where typically the vast majority of instances are unlabeled: $N \gg L$.
%
%
In addition, we are given a set of $K$ subsets of $\mathcal{X}$, which we call \emph{bags}:
$$ \mathcal{B} = \{ \mathcal{B}_1, \mathcal{B}_2, \ldots \mathcal{B}_K\} , \mathcal{B}_k \subseteq \mathcal{X}_N \cup \mathcal{X}_L.$$  For example, $\mathcal{B}_1$ could be the group of mobile users who tend to wake early in the morning, while $\mathcal{B}_2$ could be those users who stay up late at night. Note that bags $\mathcal{B}$ may overlap, and do not have to cover all training instances $\mathcal{X}_N$.

Finally, we have \emph{constraints} associated with the labels within bags ($\{y^{*}_i : i \in \mathcal{B}_k\}$). For example, we might have rough bounds on the label average in some bag, or know that the average in one bag is higher than in another. 

We are especially interested in the case where very little information is known: constraints are \emph{loose}, and specified only for a small subset of bags. Given this information during training, our goal is to learn a function $f(\mathbf{x})$ that predicts a label for individual instances, including instances that do not have an associated bag. In the following, we discuss the two most common settings -- classification and regression. 

\subsection{Background: Ballpark Classification}
\label{subsec:Classification}

The Ballpark setting was first proposed in \cite{hope2016ballpark} for \emph{binary classification} problems. For completeness, we briefly review the classification setting here. In the next section, we extend the framework to \emph{regression} problems with new methods and properties.

In the classification setting, label space $\mathcal{Y}$ is discrete,
$y^{*}_i \in \{-1,1\}$. Let $p_k$ be the proportion of positive-labeled instances in bag $\mathcal{B}_k$:
\begin{equation}
\label{eq:p_def}
p_k = |\{i:i \in \mathcal{B}_k, y^{*}_i  =1\}/ {|\mathcal{B}_k|}
\end{equation}
(where $y^{*}_i$ is replaced with $y_i$ for instances $\mathbf{x}_i \in \mathcal{X}_L$). 

Importantly, $p_k$ is not assumed to be known (unlike related work on learning from label proportions \cite{quadrianto2009estimating}; see Section \ref{sec:RelatedWork}). Rather, the model is given weaker prior knowledge, in the form of {\bf constraints} on proportions. In \cite{hope2016ballpark}, constraints included:
\begin{compactitem}
	\item \textbf{Lower, upper bounds} on bag proportions:  $l_k \leq p_k \leq u_k,$
	\item \textbf{Differences}:  $l_{k_{12}} \leq p_{k_1} - p_{k_2} \leq u_{k_{12}}.$ 
	\remove{\item \textbf{Differences (multiplicative)}:  $l_{k_{12}} \leq \frac{p_{k_1}}{p_{k_2}} \leq u_{k_{12}}.$} 	
\end{compactitem}

For example, suppose we would like to predict whether a user is over age $65$. We may have a bag of users $\mathcal{B}_k$ who often go out at night (based on GPS readings). From prior socio-demographic research, we could know that $p_k$ is somewhere between $l_k=0.1$ and $u_k=0.3$. We may also have knowledge about the difference between users with high and low outdoor nighttime activity levels.

Our {\bf prediction function}  is $f(\mathbf{x}) = \text{sign}(\mathbf{w}^{T}\varphi(\mathbf{x}))$, where $\mathbf{w}$ is a weight vector we estimate and  $\varphi(\cdot)$ is a feature map (to simplify notation bias term $\mathbf{b}$ is dropped by assuming a vector $\mathbf{1}_{N+L}$ is appended to the features). To attain the classification goal, we use a maximum-margin approach,  formulating Ballpark classification as a \emph{bi-convex} optimization problem and solving it with an alternating minimization algorithm. For more details, see \cite{hope2016ballpark}.

\subsection{Ballpark Regression}

\label{subsec:Regression}
Many real-world problems of interest involve a continuous target, which poses a special challenge for crowdsourcing \cite{whalen2015using}. In this section, we thus extend the Ballpark setting to regression. 

Denote $\bar{y}_k  =  \frac{\sum_{i \in \mathcal{B}_{k}} y^{*}_i}{|\mathcal{B}_{k}|}$, where $y^{*}_i \in \mathbb{R}$. Similarly to the proportion constraints in the classification scenario, our constraints are over the bag averages $\bar{y}_k$. We allow constraints of the following form:
\begin{compactitem}
	\item \textbf{Lower, upper bounds} on bag averages:  $l_k \leq \bar{y}_k \leq u_k,$
	\item \textbf{Differences (additive)}:  $l_{k_{12}} \leq \bar{y}_{k_1} - \bar{y}_{k_2} \leq u_{k_{12}},$ 
	\item \textbf{Differences (multiplicative)}:  $l_{k_{12}} \leq \frac{\bar{y}_{k_1}}{\bar{y}_{k_2}} \leq u_{k_{12}}.$ 	
\end{compactitem}
 We extend \cite{hope2016ballpark} and incorporate multiplicative differences, as these often may be intuitive for crowd workers (see Section \ref{subsec:recid}).

 Our goal is to predict a target for each $\mathbf{x}_i$ using a {\bf regression function} $f(\mathbf{x}) =\mathbf{w}^{T}\varphi(\mathbf{x})$. We directly model the latent variable $\mathbf{y}^{*}$ -- the vector of unknown labels $y^{*}_i \in \mathbb{R}$ -- in a constrained optimization problem. Let $\mathcal{R}$ be the subset of $\mathcal{B}$ for which we have upper and/or lower bounds. Let $\mathcal{D}$ be the set of tuples $(\mathcal{B}_{k_1},\mathcal{B}_{k_2})$ for which we have difference bounds. We formulate the following convex optimization problem:
\begin{equation}
\label{eq:prob_conv_reg}
\begin{aligned}
\underset{\mathbf{y,w}}{\text{argmin}} & \frac{1}{2}  \mathbf{w}^T\mathbf{w} +  \frac{C_N}{N} \sum\limits_{i=1}^{N} || y_i - \mathbf{w}^{T}\varphi(\mathbf{x}_i))||^2_2  \\
& + \frac{C_L}{L}\sum\limits_{j=N+1}^{N+L} || y_j - \mathbf{w}^{T}\varphi(\mathbf{x}_j))||^2_2 \\
 s.t. \quad 
& l_k \leq  \hat{\bar{y}}_k \leq u_k \quad  \forall \{k : \mathcal{B}_{k} \in \mathcal{R} \} \\
& l_{k_{12}} \leq \hat{\bar{y}}_{k_1}-\hat{\bar{y}}_{k_2} \leq u_{k_{12}} \quad
\forall \{k_1 \neq k_2 : (\mathcal{B}_{k_1},\mathcal{B}_{k_2}) \in \mathcal{D} \}
\end{aligned}
\end{equation}

where $\hat{\bar{y}}_k  =  \frac{\sum_{i \in \mathcal{B}_{k}} y_i}{|\mathcal{B}_{k}|}$ is the estimated mean in bag $\mathcal{B}_{k}$, $l_k$ (or $u_k$) can be $-\infty$  ($+\infty$) if not given as input, and analogously for difference bounds $l_{k_{12}} (u_{k_{12}})$; we omit multiplicative difference bounds for brevity. $C_N$ and $C_L$ are cost hyperparameters for unlabeled and labeled instances, respectively. $C_L$ controls the weight we give to labeled instances versus prior knowledge on $\mathcal{B}$.

Importantly, unlike in Ballpark classification, this problem is convex and thus we are guaranteed to find its global minimum. We are also able to derive some  insights into the regression model. Problem \ref{eq:prob_conv_reg} is quadratic with respect to $\mathbf{y}$, allowing the use of dedicated solvers. Furthermore, we note that with respect to $\mathbf{w}$, we can write out the minimizer (as function of $\mathbf{y}$) explicitly, as follows.

\begin{equation}
\label{eq:min_w}
\begin{aligned}
\underset{\mathbf{w}}{\text{argmin}} & \frac{1}{2}  \mathbf{w}^T\mathbf{w} +  \frac{C_N}{N} \sum\limits_{i=1}^{N} || y_i - \mathbf{w}^{T}\varphi(\mathbf{x}_i))||^2_2 \\
& + \frac{C_L}{L}\sum\limits_{j=N+1}^{N+L} || y_j - \mathbf{w}^{T}\varphi(\mathbf{x}_j))||^2_2\\
\end{aligned}
\end{equation}
Denote the solution to  Problem \ref{eq:min_w} as $\mathbf{w}^*$. \remove{Denote the feature matrices for all instances ${1,\ldots,N}$ and ${N+1,\ldots,N+L}$ as $\mathbf{\Phi}_1$ and $\mathbf{\Phi}_2$, respectively. Correspondingly, denote the latent labels ${\mathbf{y}_1,\ldots,\mathbf{y}_N}$ and observed labels ${\mathbf{y}_{N+1},\ldots,\mathbf{y}_{N+L}}$ as $\mathbf{y}_1$, $\mathbf{y}_2$.}  As is readily seen, this is essentially a (weighted) ridge regression problem. In this paper, we are interested in the case where we have no labels at all. In our experiments we do not use any labeled instances, thus at this point we set $C_L = 0$ for ease of exposition, and recover the familiar ridge solution with closed-form solution w.r.t $\mathbf{y}$, $(\lambda\mathbf{I} + \mathbf{\Phi}^T\mathbf{\Phi})^{-1}\mathbf{\Phi}^T\mathbf{y}$,
\remove{
\begin{equation}
\label{eq:opt_w}
\begin{aligned}
\mathbf{w}_* = \frac{1}{N}(\mathbf{I} + \frac{1}{N}(C_N\mathbf{\Phi}_1^T\mathbf{\Phi}_x +C_L\mathbf{\Phi}_2^T\mathbf{\Phi}_2))^{-1}(C_N\mathbf{\Phi}_1^T\mathbf{y}_1 + C_L\mathbf{\Phi}_2^T\mathbf{y}_2)
\end{aligned}
\end{equation}.
}
%
%

\noindent where $\mathbf{\Phi}$ is the feature matrix for instances $\{1,\ldots,N\}$ and $\lambda$ is a regularization hyperparameter (inversely proportional to $C_N$).  Now, we can plug the expression for $\mathbf{w}_*$ back into the objective function, and solve a quadratic program for $\mathbf{y}$:
\begin{equation}
\label{eq:solve_y}
\begin{aligned}
\underset{\mathbf{y}}{\text{argmin}} & \frac{1}{N} \sum\limits_{i=1}^{N} || y_i - \mathbf{w}_*^{T}\varphi(\mathbf{x}_i))||^2_2\\
s.t. \quad 
& l_k \leq  \hat{\bar{y}}_k \leq u_k \quad  \forall \{k : \mathcal{B}_{k} \in \mathcal{R} \} \\
& l_{k_{12}} \leq \hat{\bar{y}}_{k_1}-\hat{\bar{y}}_{k_2} \leq u_{k_{12}} \quad
\forall \{k_1 \neq k_2 : (\mathcal{B}_{k_1},\mathcal{B}_{k_2}) \in \mathcal{D} \},
\end{aligned}
\end{equation}

which yields our final, optimal weight vector $\mathbf{w}_*$. Intuitively, the first ``step'' (expressing $\mathbf{w}_*$ w.r.t $\mathbf{y}$) finds a weight vector $\mathbf{w}$ predicting $\mathbf{y}$, and linear constraints on $\hat{\bar{y}}_k $ ensure that we find an assignment to $\mathbf{y}$ that satisfies constraints given for bags $\mathcal{B}_k$.

In some cases we may have constraints that apply \textit{globally} -- such as that $y_i \ge 0  \forall i$, or that the global target mean  is within a certain range. These constraints can easily be incorporated by setting the appropriate $\mathcal{B}_{k}$ (e.g., one bag consisting of the entire training set). In Section \ref{sec:CrowdRelated}, we discuss how to use crowds to build constraints.

\xhdr{Learning as a feasibility problem} We further develop an alternative method for Ballpark regression. Here, we do not optimize for the latent $\mathbf{y}$ and $\mathbf{w}$ concurrently, but only for $\mathbf{w}$ in a feasibility optimization problem. Noting that our quantity of interest $\hat{\bar{y}}_k  =  \frac{\sum_{i \in \mathcal{B}_{k}} y_i}{|\mathcal{B}_{k}|}$ can be modeled as $\frac{\sum_{i \in \mathcal{B}_{k}} \mathbf{w}^{T}\varphi(\mathbf{x_i})}{|\mathcal{B}_{k}|}$ under the standard linear regression premise that the (conditional) expected value of the target is a linear function of the input, we solve the following simple and intuitive convex program:
\begin{equation}
\label{eq:prob_init_reg}
\begin{aligned}
\underset{\mathbf{w}}{\text{argmin}} & \frac{1}{2}  \mathbf{w}^T\mathbf{w} + \frac{C_L}{L}\sum\limits_{j=N+1}^{N+L} || y_j - \mathbf{w}^{T}\varphi(\mathbf{x}_j))||^2_2 \\
s.t. \quad 
& l_k \leq  \frac{\sum_{i \in \mathcal{B}_{k}} \mathbf{w}^{T}\varphi(\mathbf{x_i})}{|\mathcal{B}_{k}|} \leq u_k \quad  \forall \{k : \mathcal{B}_{k} \in \mathcal{R} \} \\
& l_{k_{12}} \leq \frac{\sum_{i \in \mathcal{B}_{k_1}} \mathbf{w}^{T}\varphi(\mathbf{x_i})}{|\mathcal{B}_{k_1}|}-\frac{\sum_{i \in \mathcal{B}_{k_2}} \mathbf{w}^{T}\varphi(\mathbf{x_i})}{|\mathcal{B}_{k_2}|} \leq u_{k_{12}} \\ 
& \quad \forall \{k_1 \neq k_2 : (\mathcal{B}_{k_1},\mathcal{B}_{k_2}) \in \mathcal{D} \} \\
\end{aligned}
\end{equation}

Thus, in Problem \ref{eq:solve_y} we are essentially finding an assignment $\hat{\mathbf{y}}$ to $\mathbf{y}$, such that $\hat{\mathbf{y}}$ satisfies bag constraints and is at the same time close (in $l_2$ norm) to the ordinary (unconstrained) ridge prediction based on $\mathbf{\Phi}, \hat{\mathbf{y}}$. In contrast, in the feasibility problem, we find a regularized solution $\mathbf{w}$ such that \textit{predictions themselves} satisfy the Ballpark constraints. The feasibility problem, which does not attempt to find an assignment $\hat{\mathbf{y}}$, clearly has less parameters and is thus lighter and runs blazingly fast in our experiments (less than a second for all the experiments we describe in the paper). We observed in our experiments that the two approaches led to similar results (with a slight advantage to the former that also adjusts $\mathbf{y}$), and thus we focus on using the solution to Problem \ref{eq:prob_conv_reg} for simplicity.\footnote{Code and more materials can be found at: \url{https://github.com/ttthhh/ballpark.git}
}

\xhdr{PAC formulation and sample complexity bound} We briefly derive a basic sample complexity bound for Problem \ref{eq:prob_init_reg} 
using the general PAC (Probably Approximately Correct) framework. To simplify, we set $C_L =0$ (no labeled data), and derive a dimension-based bound rather than norm-based by ignoring regularization term $\frac{1}{2}  \mathbf{w}^T\mathbf{w}$. Keeping this term would require some slightly more technically involved analysis. Our main goal here is just to get some preliminary and intuitive theoretical grounding of the model.

\begin{lemma}
For $\mathbf{w} \in \mathbb{R}^d$, and under the modification mentioned above, Problem 
\ref{eq:prob_init_reg} can be cast in the general PAC learning model, and  sample complexity $m_{\mathcal{H}}$ is bounded by $\mathcal{O}(\frac{d}{\epsilon^2})$, where $\mathcal{H}$ is a hypothesis class $\mathbf{w}$ induces over instances $\mathbf{x}$ and bags $\mathcal{B}$, and $\epsilon$ follows the standard PAC notation (see \cite{shalev2014understanding}).  
\end{lemma}	

\remove{
\begin{lemma}
For $\mathbf{w} \in \mathbb{R}^d$, and under the modification mentioned above, Problem 
\ref{eq:prob_init_reg} can be cast in the general PAC learning model, and we obtain the basic sample complexity bound of 
\begin{equation}
\begin{aligned}
m_{\mathcal{H}} (\epsilon,\delta) \leq \frac{64d + 2\log(\frac{2}{\delta})}{\epsilon^2}, 
\end{aligned}
\end{equation}
where $\mathcal{H}$ is an hypothesis class $\mathbf{w}$ induces over instances $\mathbf{x}$ and bags $\mathcal{B}$, and $\epsilon,\delta$ follow the standard PAC notation (see \cite{shalev2014understanding}).  
\end{lemma}	
}
\textit{Proof sketch.} We incorporate the Ballpark constraints into a $0$-$1$ loss function, taking the value $1$ when constraints are violated and $0$ otherwise. Parameterizing  a hypothesis class $\mathcal{H}$ over instances and bags with $\mathbf{w}$, we obtain a PAC formulation. We derive a dimension-based bound using the practical discretization trick to treat $\mathcal{H}$ as finite, applying corollary 4.6 in \cite{shalev2014understanding} to obtain the bound. $\square$

\xhdr{Optimizing hyperparameters}
\label{sec:optimC}
In practice, we need to tune hyperparameter $C_N$ ($\lambda$). This is typically done with cross-validation (CV) grid search. However, standard CV is impossible here as we have no labels to compute accuracy on held-out data. 

We thus use a variant of CV called Constraint Violation Cross Validation (CVCV) developed in \cite{hope2016ballpark} for the classification case, which is readily adaptable to regression. We run $K$-fold CV, splitting each bag $\mathcal{B}_{k}$ into training and held-out subsets. The intuition is that the label average in uniformly-sampled subsets of a bag is similar to $\bar{y}_k$ in the entire bag.
For each split, we solve Problem \ref{eq:prob_conv_reg} on training bags, and then compute by how much constraints are violated on \textit{held-out} bags. More formally, we compute the average 
deviations from bounds, $max(\hat{\bar{y}}_k -u_k,0)$, $max(l_k-\hat{\bar{y}}_k,0)$ for $\hat{\bar{y}}_k$ the estimated label mean in the held-out subset of bag $k$. We do so over a grid, and select the $C_N$ ($\lambda$) with lowest average violation. This simple approach leads to good hyperparameter selections in practice.

\section{Ballpark with Crowd Constraints} 
\label{sec:CrowdRelated}
In the previous sections we presented the Ballpark formulation. Most importantly, in Ballpark learning we do not assume to be given labels, but rather weak information in the form of constraints. A natural question is how we obtain these constraints.

In \cite{hope2016ballpark} we assumed that constraints came from experts or some other source of domain knowledge. It was left as an open question whether the methods would still be effective using noisy information from a crowd of non-experts. Obtaining constraints from the crowd raises some interesting points which we discuss below.

\xhdr{Constraint aggregation}
There are multiple ways to build constraints by querying the crowd. Importantly, virtually all of these methods require \emph{aggregation} of crowd guesses. 
A large body of work in crowdsourcing for machine learning and ``wisdom of the crowds'' (WoC) deals with aggregating human guesses \cite{lorenz2011social,venanzi2014community,gao2015truth,prelec2017solution,ugander2015wisdom, whalen2015using}. The objective of these methods is typically to obtain accurate point estimates of some quantity. 
A simple and popular approach is to take the mean of guesses, potentially weighted by worker quality. More elaborate methods construct rich probabilistic models \cite{venanzi2014community,whalen2015using,gao2015truth} to capture different worker properties (e.g., systematic biases). \remove{These models require, of course, algorithms and expertise to estimate parameters and obtain refined estimates.} Other methods rely on multiple guesses and incentives \cite{ugander2015wisdom} or assume labeled ground truth \cite{whalen2015using}, requiring more resources and a more taxing experience for workers and practitioners.
Importantly, rather than aiming to obtain a point estimate, in the Ballpark framework we only require \emph{broad intervals} ``bracketing'' the true label average of bags. Instead of focusing  efforts on aggregating crowd guesses ``accurately'', we rely on a machine learning model to use these intervals and predict accurate individual labels. 

Thus, one simple method that is suitable for the Ballpark framework is to elicit crowd guesses on group label averages, and then construct bag bounds based on percentiles of the empirical guess distribution (see Section \ref{sec:Experiments2}).
\remove{For example, in experiments below we ask workers to guess average  prices for certain groups of apartments. We build lower and upper bounds and for the average price in bag by simply taking the $q$ and $1-q$ percentiles of answers (see Section \ref{sec:Experiments1} for extensive sensitivity analysis, showing overall robustness to $q$).} An alternative, less common approach in WoC is to forgo point estimates altogether and consider the distribution of all possible values, such as assigning different probabilities to different \emph{intervals} \cite{haran2010simple}. In \cite{park2015aggregating} multiple interval guesses from a crowd are aggregated to improve predictions. Loosely inspired by this, we also consider asking people to \emph{guess intervals} bracketing label averages in groups and simply take the means of the upper and lower bounds (see Section \ref{sec:CrowdAirbnb}).

\remove{ As we show in our experiments, this approach is able to get good results, with the advantage of freeing the practitioner of having to select $q$.} 

\remove{ In \cite{haran2010simple}, people were asked to estimate the temperature in Pittsburgh one month into the future, by assigning probabilities to $9$ temperature intervals. Here, the aim is not to obtain point estimates -- intervals and their associated probabilities are the end-game. In small niche of WoC, multiple interval guesses from a crowd are aggregated to improve prediction quality\cite{park2015aggregating}.}

\xhdr{Constraint feasibility}  In the Ballpark problem formulation we impose hard constraints on label means. In practice, we may face certain constraints that are infeasible. This could happen, for example, when we receive misspecified lower and upper bounds from non-experts. As we demonstrate in Section \ref{sec:CrowdBoston}, constraints can be made soft by adding slack variables $\xi$ to the infeasible subset. \remove{We discuss and demonstrate the effects of constraints and different strategies in more detail in the Section \dnote{ref}.}

\xhdr{Tasks Suitable for Crowd Constraints}
We believe the intuition of crowds could be especially useful 
when trying to learn hidden behavioral, sociological or commercial attributes. In our experiments, we examine two predictive tasks: Predicting return to jail (complex behavioral property) and predicting rental prices (commercial intuition).
Our method is less suitable for cases where the bias of crowdworkers is so extreme as to \textit{substantially} skew even the lower or upper bounds on label averages. This could be caused by domain ignorance or deeply ingrained beliefs. 



\label{sec:Regression}
\section{Evaluation (1): Synthetic constraints} 
\label{sec:Experiments1}
For our first evaluation task, we wish to explore the robustness of Ballpark Learning to different constraint settings, examine the effect of constraints on prediction accuracy, and compare Ballpark to the supervised setting. 
We create artificial bags based on a real dataset, varying bag and constraint construction to illustrate some of the different factors that come into play. As the classification setting was evaluated in \cite{hope2016ballpark}, we focus on regression.

\remove{
For our first evaluation task, we set out to explore {\bf when does the Ballpark setting work, and how
well?} We are interested to learn how well Ballpark compares to supervised learning, explore the robustness of ballpark regression to different constraints settings, and examine the effect of constraints on prediction accuracy.Note that our goal is not to compete with the host of methods available for classification and regression in terms
of accuracy, but rather to find out how far we can get with very little information available.
}
\remove{
In \cite{hope2016ballpark}, an empirical exploration of the effects of different regimes of constraints is conducted for the classification setting. The authors create artificial bags based on a real data set, and vary the tightness of constraints, reporting classification accuracy and illustrating some of the different factors that come into play. In this section we extend these experiments to regression. }

\xhdr{Data} The Boston dataset is a well-known, small dataset ($506$ instances) available from the StatLib repository \cite{vlachos2000statlib} that is popular for evaluating regression models. It contains information on Boston housing characteristics and values. The target variable is the median value of owner-occupied homes in different areas. Each area has $13$ features, including crime rates, air pollution (nitric oxides concentration), average number of rooms, and more. We estimate the prices of apartments using ballpark regression with bags based on crime levels, pollution, and number of rooms. 

\xhdr{Constraints} We construct bags based on only three variables: crime (CRIM), pollution (NOX), and average number of rooms (RM). We discretize these variables into three bags each (cutoffs at the $0.33$ and $0.66$ percentiles). This yields $9$ bags, such as $\mathcal{B}_\text{high crime}$, $\mathcal{B}_\text{medium crime}$, $\mathcal{B}_\text{low crime}$. Next, we compute (true) bag averages, and construct pairwise constraints based on their partial ordering. Bounds on bag means are built by introducing a multiplicative error term multiplying the true bag means ($10\%$ in each direction, with apartment prices measured in units of \$$1000$). If for bag $\mathcal{B}_k$ the true bag mean is $\bar{y}_k$, then upper and lower bounds are $(1+\epsilon)\bar{y}_k,(1-\epsilon)\bar{y}_k$, respectively, where $\epsilon (=0.1)$ is the error term for individual bags. Bounds on bag differences are created in the same way.

\begin{figure*}[!htb]
    \centering
    \begin{subfigure}{0.33\textwidth}\includegraphics[ height=3.9cm]{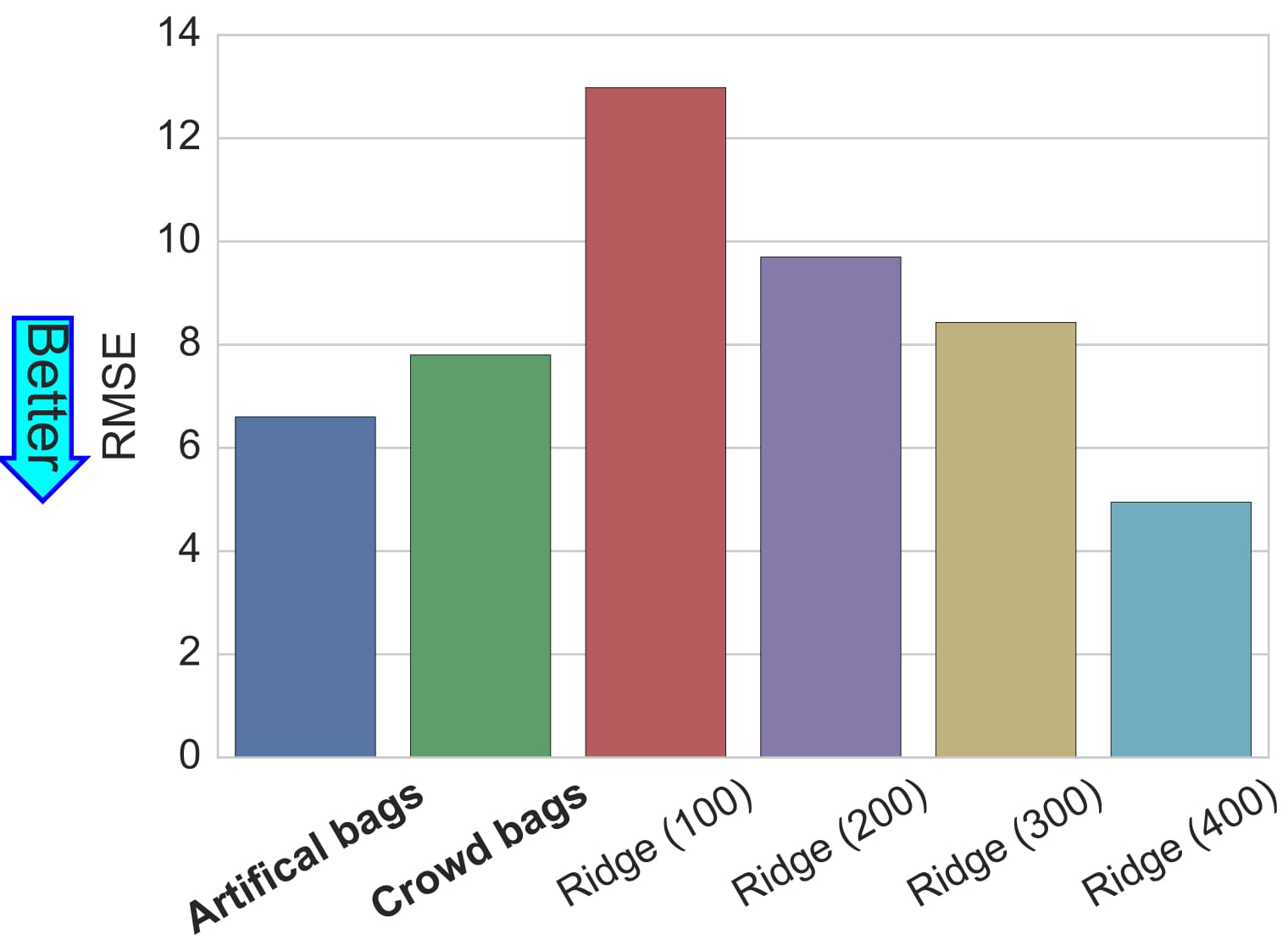}
    \caption{}
    \end{subfigure}%
    ~ 
    \begin{subfigure}{0.33\textwidth}\includegraphics[height=4cm]{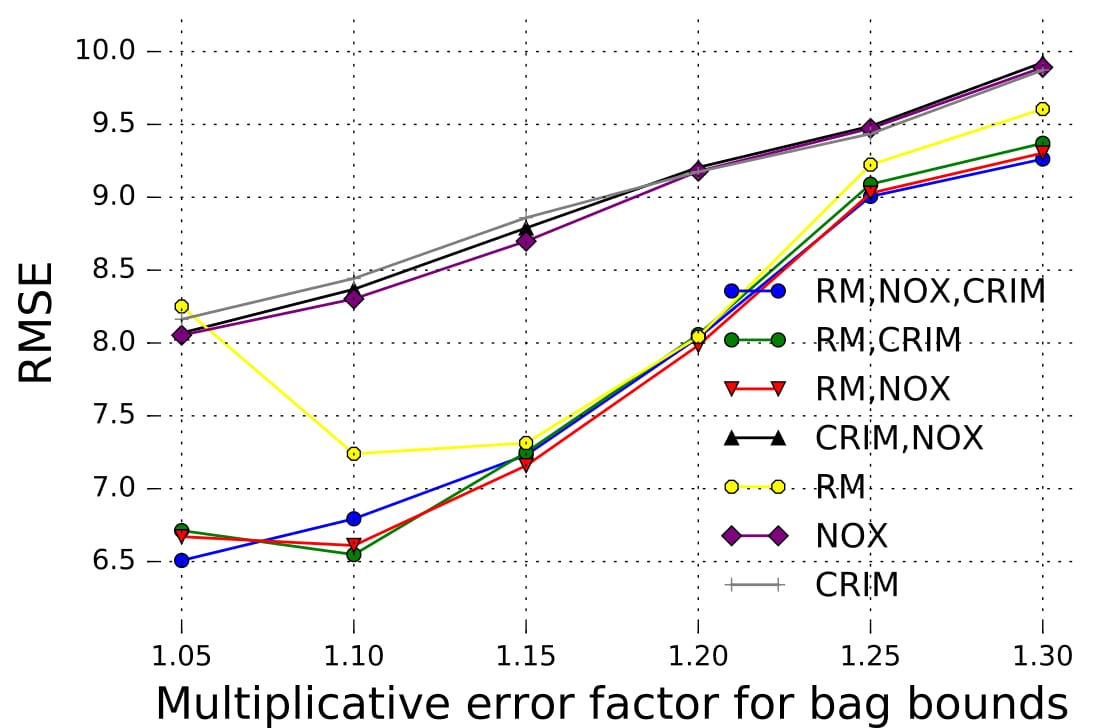}
     \caption{}
    \end{subfigure}%
    ~ 
    \begin{subfigure}{0.33\textwidth}\includegraphics[ height=4cm]{{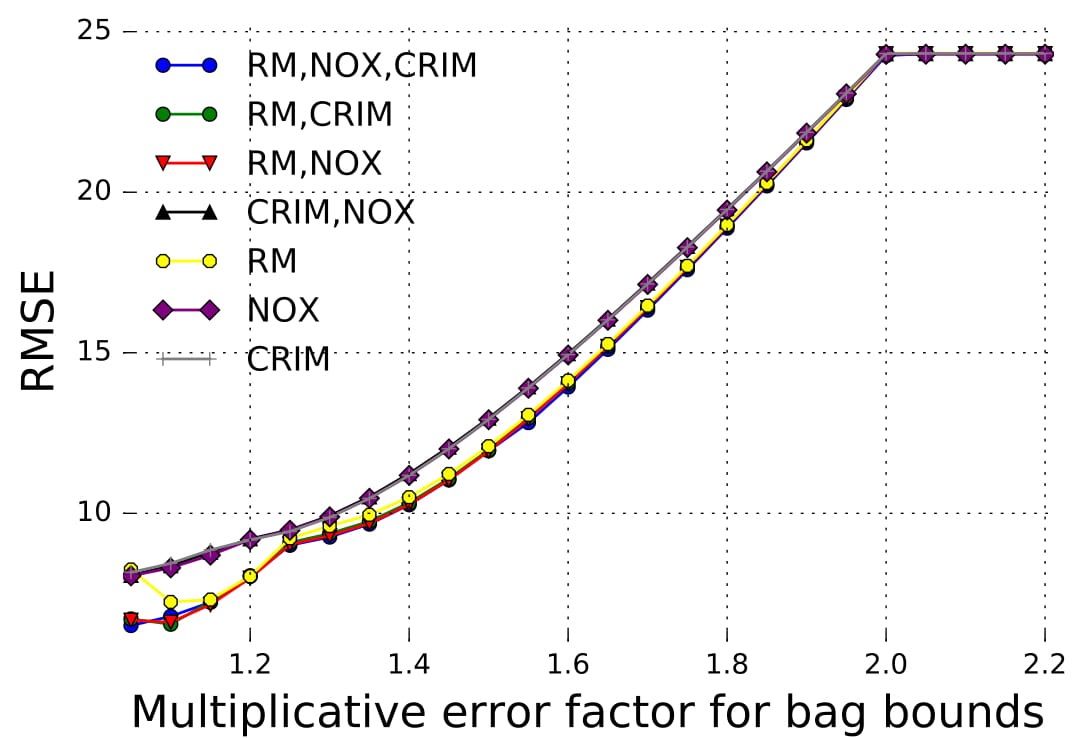}}
     \caption{}
    \end{subfigure}%
     \vspace{-2mm}
    \caption{\small  (a) Boston accuracy. Ballpark surpasses or rivals ridge regression with a considerable amount of ground-truth labels. As expected, constraints artificially constructed assuming more prior knowledge, are better than constraints aggregated from crowd guesses (Section \ref{sec:CrowdBoston}).  (b) Varying multiplicative error term on synthetically-built constraints. (c) Continuing plot (b) for extremely large error terms \label{fig:boston_res}}
\end{figure*}

We compare our method to supervised ridge regression with an increasing number of labeled examples, reporting average RMSE results (over $5$-fold CV). Our goal is not to compete with the host of methods tested on this benchmark, but rather to see how far we can get using only weak information. As seen in Figure \ref{fig:boston_res}(a), by using weak domain knowledge on bags we are able to either surpass or rival ridge regression with a considerable amount of ground-truth labels. With $300$ labels (accounting for $\approx 60\%$ of the entire data), there is still a considerable gap in RMSE in favor of ballpark regression.
We select regularization hyperparameter with the CV method described Section \ref{sec:Formulation}. Due to the small size of this dataset, we do so with only $3$ inner training folds, which in our experiments was enough to reach good results.

\xhdr{Sensitivity analysis}
\remove{Our methods assume constraints that may be broad, but still reflect the true underlying distribution. These constraints may come from experts, from prior knowledge, literature, or from crowdsourcing.} 
We now explore what effect quality of constraints has on prediction quality. For the synthetic constraints described above, we vary the multiplicative factor $\epsilon$, gradually loosening upper and lower bounds on bag means. As seen in Figures \ref{fig:boston_res}(b)-(c), at first error grows rather slowly with $\epsilon$, but then picks up when constraints become broad beyond reason. Indeed, assuming for instance that a true bag mean is $20$K, then $\epsilon=0.5$ means our upper and lower bounds specify the very uncertain range $[10\text{K},30\text{K}]$. Even with such a broad set of constraints obtained with $\epsilon=0.5$ we obtain error that is slightly better than using ridge regression with $100$ labels. In real applications, constraints too weak may be dropped, or more information could be collected to tighten them. 

In addition to varying $\epsilon$, we also examine the effect of bag choice. In Figures \ref{fig:boston_res}(b) and \ref{fig:boston_res}(c), we show results for using all three bags (RM+NOX+CRIM), pairs of bags (RM+NOX, RM+CRIM,NOX+CRIM), and singletons (RM,NOX,CRIM). Results are fairly robust to this choice but, as expected, using fewer bags overall leads to inferior results. The gap is more pronounced in the range of relatively tighter constraints (up to $\epsilon=0.2$). It is also evident that RM's (number of rooms) contribution to informative bag construction is strongest. Using only this variable for bag construction gives inferior results at first, but as  $\epsilon$ grows the performance of RM-only bags rapidly gets very close to richer bag constructions based on the other variables. 

\remove{
\begin{figure}

(a)\includegraphics[width=0.75\linewidth, height=4cm]{boston_bag_factor_tests_ARTIFICIAL_ZOOM} \\
        (b)\includegraphics[width=0.75\linewidth, height=4cm]{boston_bag_factor_tests_ARTIFICIAL_NEW} 

  \caption{\small \textbf{Constraint effects}. Accuracy results on cross-validation. \textbf{(a)} Varying multiplicative error term on synthetically-built constraints. \textbf{(b)} Continuing plot (a) for extremely large error terms \label{fig:const_bost}}
\end{figure}
}
\section{Evaluation (2): Crowdsourced Constraints}
\label{sec:Experiments2}
In the previous section, we evaluated the Ballpark setting with synthetic constraints. 
We now set out to discover whether we can obtain good results when constraints are crowdsourced instead.

In particular, we ask: {\bf Can crowds provide good constraints?} We are interested in both \emph{efficiency} and \emph{effectiveness}: Can we collect {highly noisy and biased} guesses from a crowd of non-experts and still obtain good instance-level predictions? Can collecting group constraints require less effort and resources than the standard practice of collecting individual labels from the crowd? How should we aggregate crowd guesses to get the best ``bang for the buck''?

\subsection{\bf Classification with Crowds: Recidivism}
\label{subsec:recid}

We start by focusing on \emph{classification} problems. 

\xhdr{Motivation and data} In the United States, a large share of crime is committed by inmates released from prison \cite{chen2007harsher}. About two-thirds of prisoners released across $30$ US states in 2005 were re-arrested within 3 years \cite{recid2010Justice}. Recently, statistical learning methods have been used for risk assessments attempting to predict the danger an offender would pose after release \cite{berk2009forecasting} to inform sentencing decisions.

\remove{Among prisoners released across $30$ US states in 2005, about two-thirds were arrested for a new crime within 3 years, and three-quarters were arrested within 5 years \cite{recid2010Justice}.Risk assessments for offenders are often used to inform the sentencing 
decisions of convicted offenders, attempting to predict the danger an offender would pose after release \cite{berk2009forecasting}. Traditionally, these assessments were given based on subjective judgments by officials, while in recent years statistical learning methods have also been used \cite{berk2009forecasting}.}

\begin{figure}[b!]
	\includegraphics[height=4.8cm]{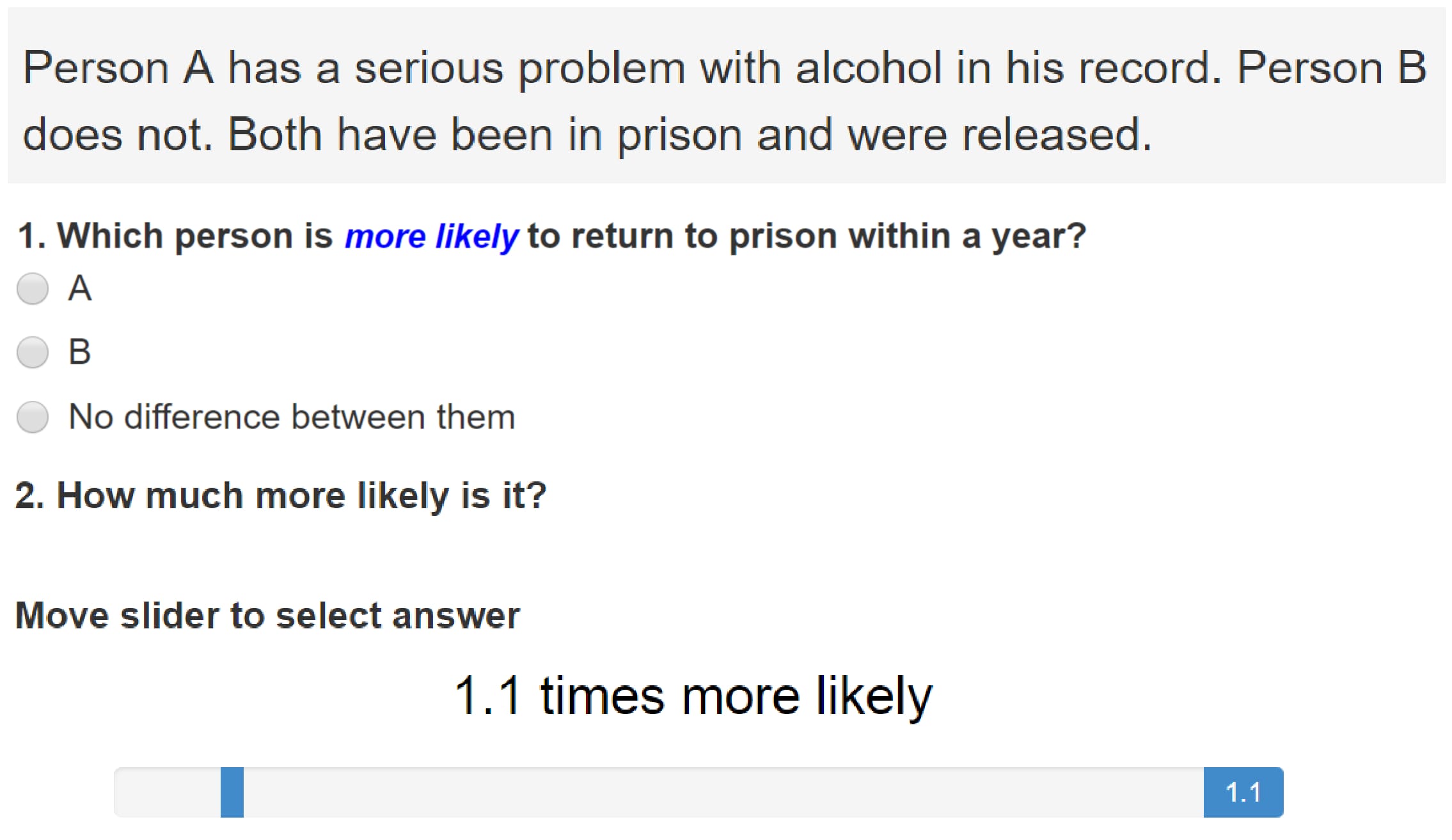}
	\caption{\small Recidivism AMT task example. Querying crowd workers for pairwise constraints on bags of instances.\label{fig:recidscreen}}
\end{figure}

We use two datasets with cohorts of inmates released in 1978 ($N=9327$) and 1980 ($N=9549$) from a North Carolina prison \cite{Schmidt84recid}. The target variable indicates whether an inmate returned to prison within a year of release. Features include race, gender, age, alcoholism, serious drug use, supervision after release, marital status, conviction due to felony/misdemeanor, and more.

\xhdr{Experiment design}
We ask Amazon Turk workers to assess the likelihood of released inmates to return to jail, based on coarse information on groups. 
We build ``bags" of inmates based on the following binary variables: gender, alcoholism, drug use, supervision, marital status, conviction due to felony/misdemeanor, crime against property/person. For each variable, we have two bags (${0,1}$). Workers are asked to determine which bag has higher recidivism rates, by how much, and guess the rate in one of the bags (See Figure \ref{fig:recidscreen} for an example; more examples are in our code repository).

We ask workers to guess which inmates are more likely to return to prison (e.g., alcoholics or not) and \textit{how much} more likely it is. Workers are also asked to guess the rate of recidivism for groups. We have $16$ groups based on $8$ binary variables, but we only ask for estimates on groups corresponding to ``positive'' values of each variable (e.g., $\text{MARRIED}=1$, $\text{MALE}=1$).
Our AMT task thus consists of $8$ HITs, each corresponding to a feature, assigning $30$ US-based workers per HIT, with approval rate greater than $97\%$ and over $500$ approved HITs, for a total cost of \$$16.80$ (including fees). In all our experiments we also tried downsampling the number of answers per HIT, retaining robust results.


\xhdr{Constraint construction}
Next, we need to aggregate the crowd's replies into a set of Ballpark constraints.
%
To construct the partial ordering $\mathcal{P}$ between pairs of bags (such as between $\mathcal{B}_\text{male},\mathcal{B}_\text{female}$), we take the majority vote, which is unequivocal across all variables but one (crimes against property). It appears the crowd's intuition conforms  with ``stereotypes'' on the relative likelihood of certain groups to commit crime. To build upper and lower bounds on bag proportions and differences, we  take the $0.75$ and $0.25$ percentiles of answers and turn them into \textit{multiplicative} constraints (other choices, such as $0.9,0.1$, led to virtually the same results; see Section \ref{sec:CrowdAirbnb} for an alternative aggregation method). We upper-bound the global proportion of recidivism at $0.4$, based on the cited statistic above on general recidivism rates, which is considered common knowledge.

\xhdr{Results} We compare our label-free method to supervised baselines. These include results previously reported for this prediction task (1978 inmate cohort), taken from \cite{huang2004learning}, all of which were obtained using all labels available during training. For the 1978 cohort data, we use the same train/test splits as the authors. We also include results we obtained ourselves by training Support Vector Machines (SVM) with different amounts of labeled examples. We vary the number of labels given to SVM to demonstrate the effect the amount of labeled data has, and compare it to our ballpark approach that uses no labeled instances. For the 1980 data, we report mean cross-validation accuracy ($10$ folds). BMP (Biased Minimax Probability Machine) is a method proposed in \cite{huang2004learning} for handling imbalanced classification tasks, reported for the 1978 cohort only.

\begin{figure}[t!]
	\includegraphics[width=0.95\linewidth,height=4.5cm]{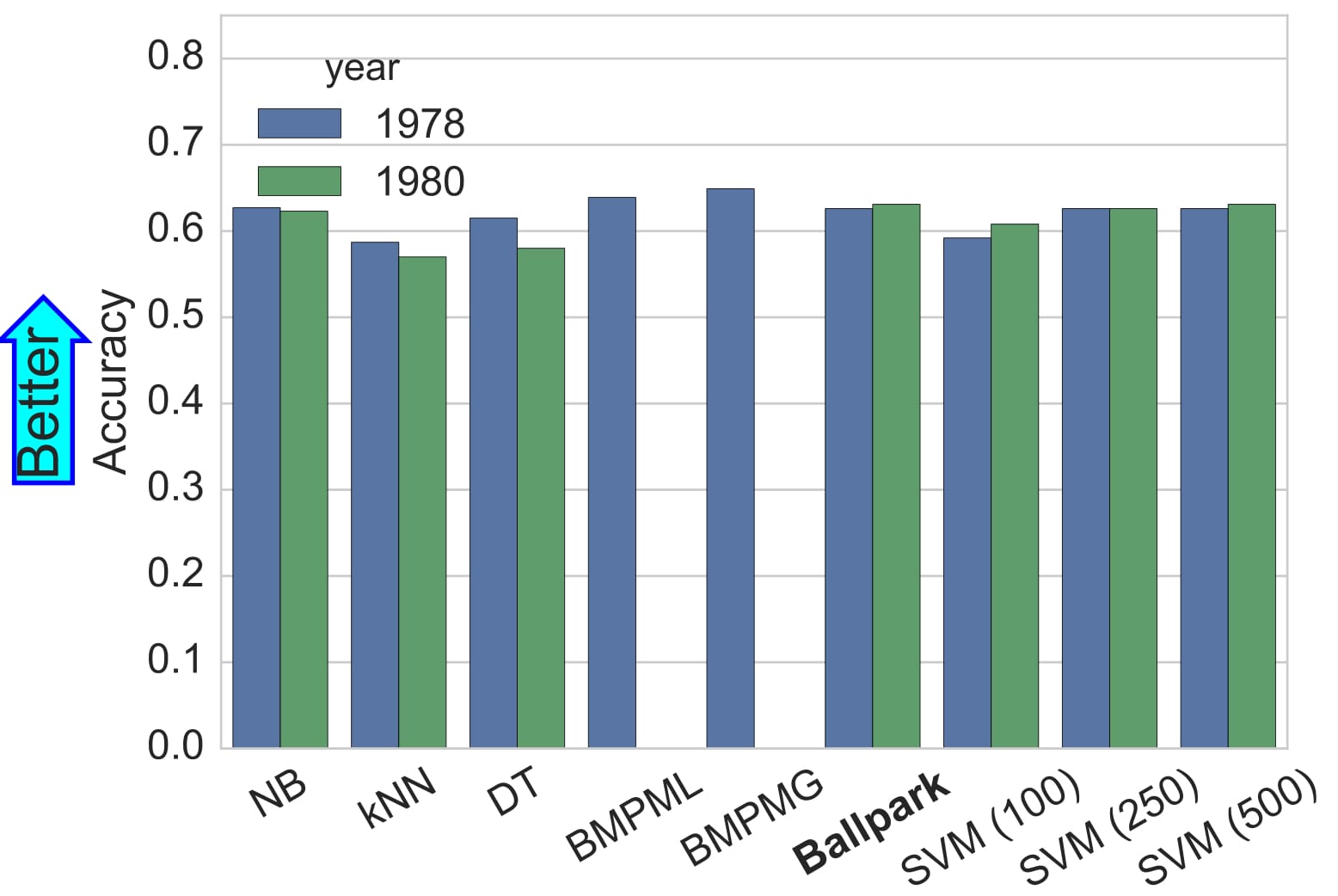} 
	\caption{\small \textbf{Recidivism results}. Comparing to baseline methods reported in \cite{huang2004learning} (1978 cohort) and trained by us (1980 cohort). BN=Naive Bayes, DT=Decision Tree, kNN=k-Nearest Neighbors, BMPML/G= \cite{huang2004learning} (all using entire labeled data), SVM = Support Vector Machines trained with an increasing number of labels.\label{fig:recid_res}}
\end{figure}

As seen in Figure \ref{fig:recid_res}, our Ballpark method achieves results that surpass or rival supervised baselines and advanced methods exposed to all \textit{true} labels, and SVM with an increasing number of labels. This is despite us not using even one ground-truth label, and leaving the construction of bag bounds to crowd workers with no real domain knowledge beyond commonplace intuition.\footnote{Incidentally, one of the workers emailed us to explain she was a retired correction officer, and ``that is why most, if not all, of my answers were negative''.} 

\begin{figure}
	\begin{subfigure}{0.5\textwidth}
		\centering
		\includegraphics[width=0.8\linewidth,height=3.6cm]{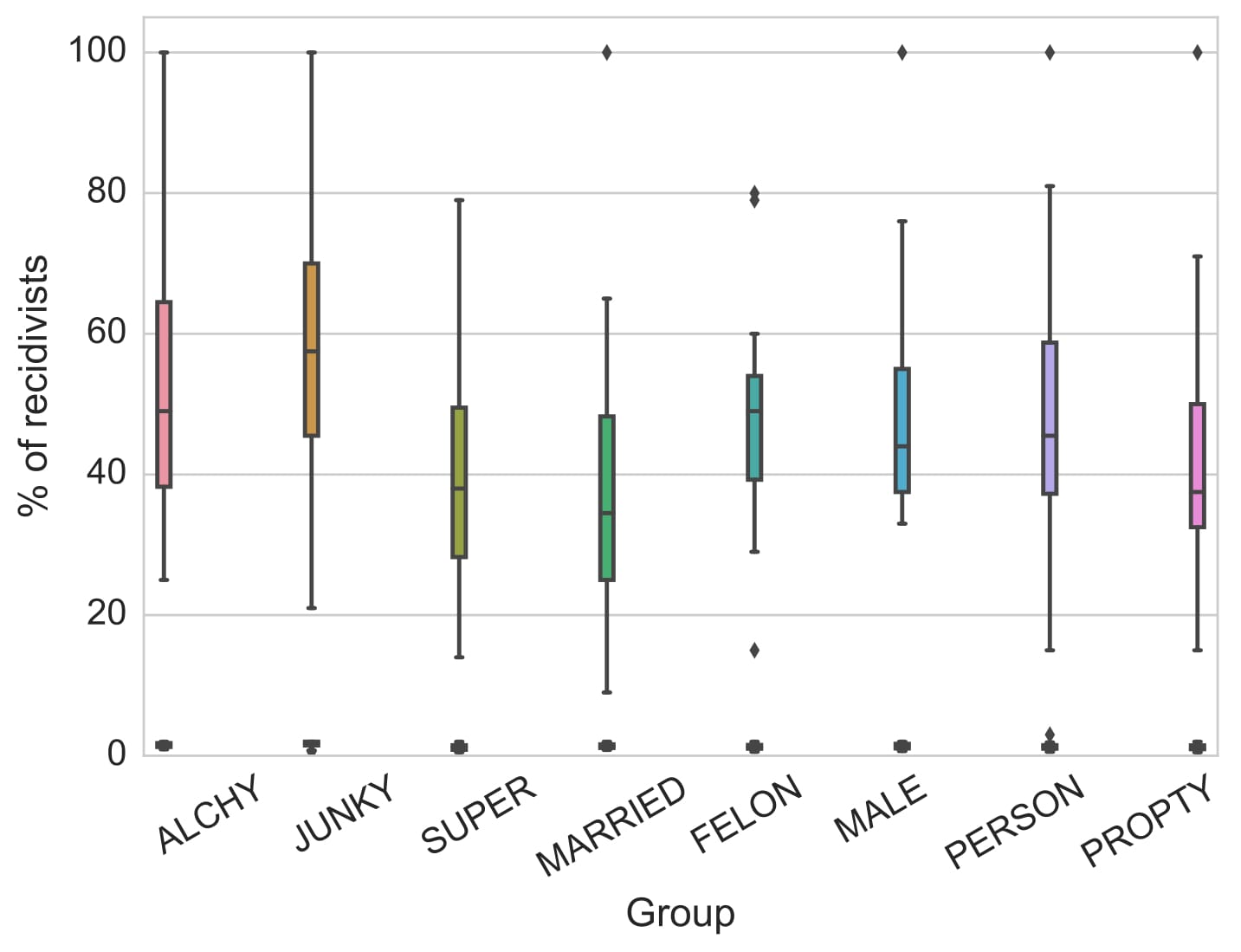} 
	\end{subfigure}
	\begin{subfigure}{0.5\textwidth}
		\centering
		\includegraphics[width=0.8\linewidth,height=3.6cm]{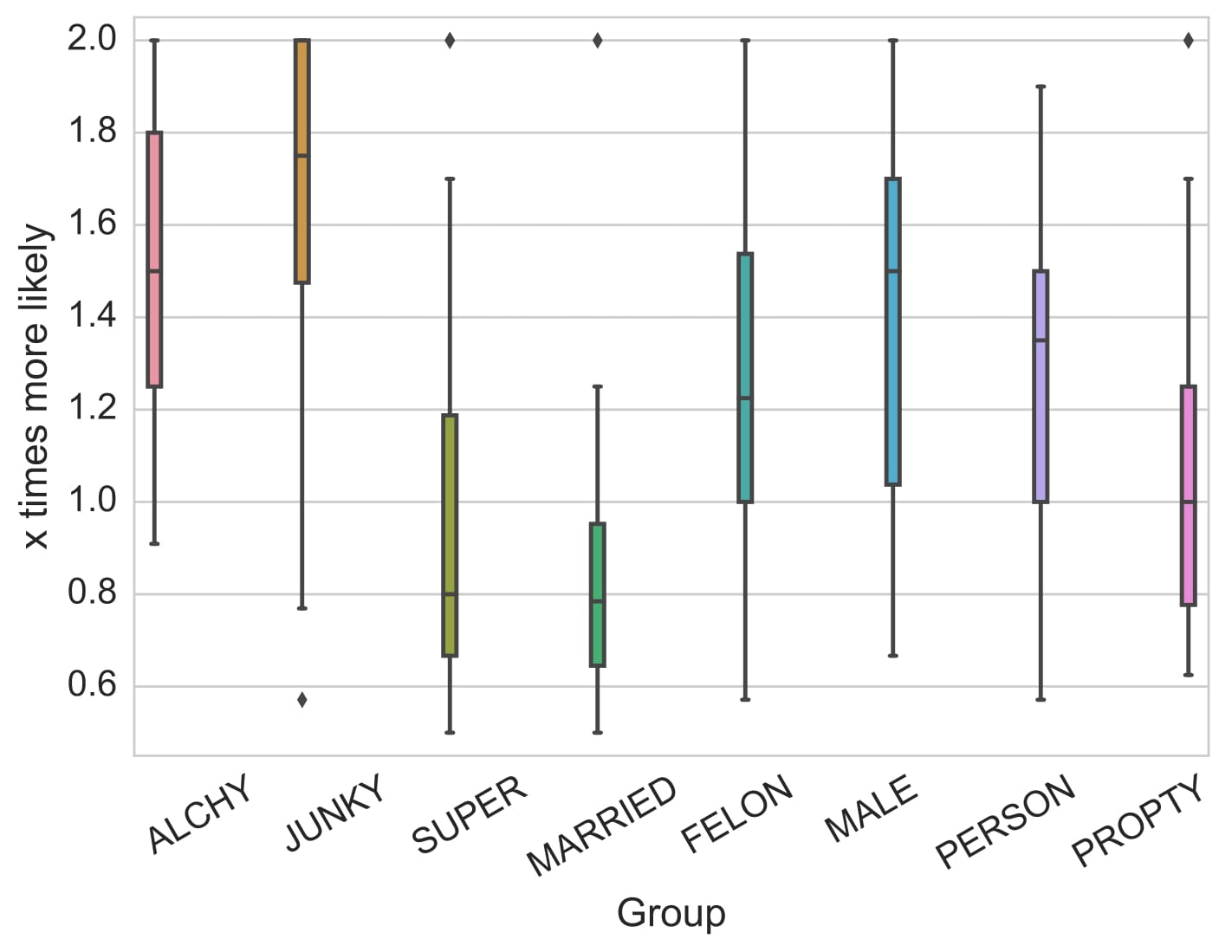} 
	\end{subfigure}
    \vspace{-6mm}
		\caption{\small Recidivism AMT worker guesses on (top) bag averages per group and (bottom) pairwise constraints.\label{fig:recid_boxplots}}
\end{figure}

\xhdr{A note on noise and bias} We observe the high amount of noise in the \textit{individual} estimates, seen in Figure \ref{fig:recid_boxplots}. In this figure, we see guesses for the percentage of recidivism per group, and for pairwise (multiplicative) differences between bags. There is large variability for both the rate of recidivism and group differences, despite giving workers basic background on the general rate of recidivism.

Note that a related line of work, learning from labels proportions \cite{quadrianto2009estimating}, assumes that true bag proportions are known. They suggest (theoretically) that sampling for bag proportions is one way to obtain accurate estimates. In practice, sampling for labels from true domain experts is typically infeasible or costly, while resorting to crowdsourcing is considered a viable option. However, while noise can potentially be averaged out, even looking at average estimates per bag leads to highly \textbf{biased} estimates, with relative errors (with respect to ground truth) of up to  $60\%$, with most errors ranging around $30\%$. These render bag average proportions highly dubious.
By using \emph{broad} constraints on averages, our methods are able to exploit crowd estimates and rival supervised methods. \remove{ and achieve results that rival supervised models based on many \emph{true} labels.}

%



\subsection{\bf Regression with Crowds (1): Boston}
\label{sec:CrowdBoston}
In our second crowd experiment, we explore Ballpark regression. We return to the Boston dataset, now using \emph{crowdsourced} constraints. We believe people should find it simpler to compare groups based on the same variable (e.g., $\mathcal{B}_\text{medium pollution},\mathcal{B}_\text{low pollution}$).
Therefore, we asked crowd workers to determine which bag has higher average apartment prices and by how much. 

To make the Boston dataset relevant in 2017, we formulate questions referring to a ``city somewhere in the world", and give some basic price statistics for this fictional city (average, minimum and maximum). Some things never change - common intuition nowadays still yields useful constraints, as our results below show.

\remove{Interestingly, the Boston data set is based on very outdated housing prices. To avoid biasing people giving us their guesses in 2017, we conceal information on location and time: We tell workers that questions refer to a ``city somewhere in the world", and give some basic statistics on prices of homes in this fictional city (average, minimum and maximum). As seen below in our obtained results, it appears that even with such coarse background, some things never change - common intuition on how homes are priced nowadays, is sufficient to provide useful constraints for this data set.}

Our AMT task thus consists of $9$ HITs, assigning $30$ US-based workers per HIT, with approval rate greater than $97\%$ and over $500$ approved HITs, for a total cost of \$$18.90$ (including fees).

We build pairwise constraints based on (clear-cut) majority votes, and construct bounds using percentiles of answers ($0.75, 0.25$ as in the recidivism task).
As seen in Figure \ref{fig:boston_res}(a), we are able to surpass supervised regression with many ground-truth labels. Unsurprisingly, synthetic constraints achieve better results due to stronger prior knowledge, but the simple bag construction from crowd constraints in this experiment still yields good results.

\begin{figure}
\includegraphics[width=0.99\linewidth,height=4.0cm]{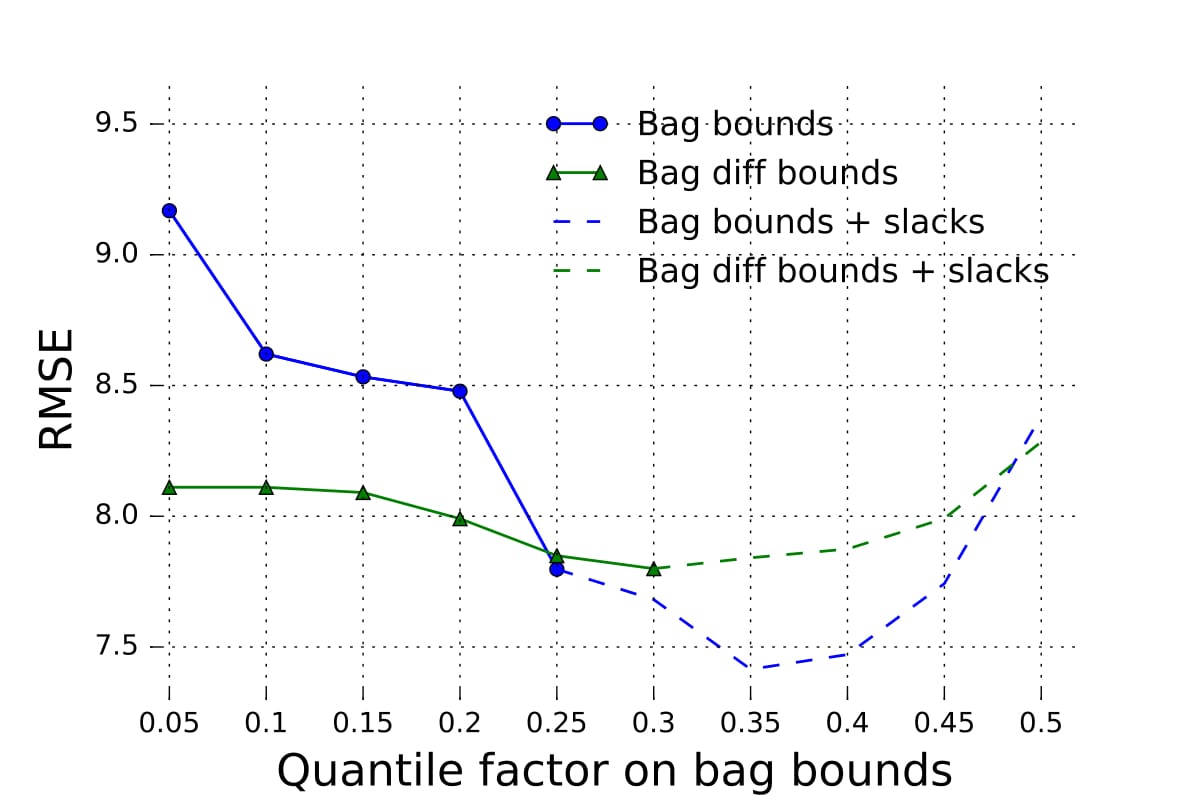} \\
  \caption{\small \textbf{Constraint effects on Boston accuracy}. Varying bound quantile on individual bags $b_l$, bag differences $d_l$, crowdsourced data. The graph stops abruptly when constraints are infeasible. We then add slack variables and continue increasing the bounds.  \label{fig:const}}
\end{figure}

\xhdr{Sensitivity analysis}
\remove{We wish to explore how the quality of constraints obtained from crowd workers affects the quality of predictions.
Note that in our crowd-sourcing experiments we aggregate workers' guesses into constraints by using a low ($0.25$) and a high ($0.75$) quantile for lower and upper bounds. We would like our results to be reasonably robust to shifting these constraints (within reason). In this section we give a brief demonstration of how the tightness of constraints could affect model performance, giving a basic illustration of some of the different factors that come into play. }
We briefly examine how the tightness of constraints could affect model performance, illustrating some of the different factors that come into play and exploring the robustness of our method.
We vary tightness of upper/lower bounds for individual bags and bag differences, reporting accuracy. We denote the lower bound percentile for individual bags as $b_l$ (e.g., $0.25$ as above), and set the upper bound to $1-b_l$. Similarly, $d_l$ and $1-d_l$ are lower and upper bound percentiles for bag differences. In Figure \ref{fig:const} we observe that after the point $b_l = 0.25$ constraints are no longer feasible, indicating badly-specified bounds. At this point we add slack variables as discussed in Section \ref{sec:CrowdRelated}. The error continues to slightly drop, and then picks up when bounds are extremely loose. 
\remove{
In our experiment, we simply increment $b_l$ until in-feasibility, and can use this as an automatic (crude) heuristic to detect bad settings, when working with crowd-provided noisy constraints. In this case, the problem is compounded with the use of a small data set. }
In the next section we show an alternative way to query crowdworkers for constraints without selecting $b_l$, $d_l$. Nonetheless, we note that even with the worst choices for parameter $b_l$ our error is still better than supervised ridge regression with $200$ labels. For parameter $d_l$, results are even more robust. 

\remove{Nonetheless, the question of how to optimally select bounds based on noisy crowd-sourced answers in our low-resource setting, and providing theoretical guarantees, is an interesting direction left to future work.}

\subsection{\bf Regression with Crowds (2): Airbnb}
\label{sec:CrowdAirbnb}
In this section we continue to compare the standard practice of collecting labels to our suggested practice of collecting group constraints.  
We also explore another simple way to obtain constraints, asking workers to \emph{guess intervals} directly.

\xhdr{Data} 
In this experiment we predict apartment prices again using a bigger, more modern apartment dataset. 
We hope workers have better intuition on apartment prices for this dataset due to its intuitive features and recency, making the evaluation more fair. 
\remove{This allows us to ask questions regarding individual instances (apartments) and using them to train supervised models.}
\remove{Airbnb is an online marketplace  enabling people to list or rent short-term lodging in residential properties. The cost of such accommodation is set by the property owner.}

Airbnb is an online marketplace enabling people to lease or rent short-term lodging. The dataset (\url{insideairbnb.com}) consists of $5147$ apartments. Our aim is to predict the price a user will enter for an apartment in Chicago in Early October 2015, based on features such as neighborhood, number of beds, amenities, and more. 
\remove{We may imagine using these predictions to automatically recommend pricing suggestions for new Airbnb users. \dnote{but airbnb has true labels}}

\xhdr{Experiment design}
We collect judgments via two separate tasks. 
%
%
First, we construct bags of apartments based solely on amenities. We look at whether or not an apartment has a \textit{TV}, a \textit{fireplace}, a building \textit{doorman} and building \textit{gym}. As in the Boston experiment, we build pairwise constraints based on majority votes by crowd workers, ask workers to guess the price difference, and construct bounds using answer percentiles. We give basic statistics on the distribution of prices in the data (average, top and bottom $5$ percentiles). 
\remove{but unlike the Boston experiment, here we expose workers to the true location and timing, hoping to better exploit their natural intuition and prior knowledge.}  Our AMT task thus consists of $4$ HITs, assigning $30$ US-based workers per HIT, with approval rate greater than $97\%$ and over $500$ approved HITs, for a total cost of \$$3.6$ (including fees).

Aside from collecting guesses on bags, we also collect hundreds of guesses on \textit{individual} instances, a standard practice for collecting labeled data via crowdsourcing.
We test the hypothesis that people are (often) better at reasoning about simple groups of instances and the \emph{pairwise ordering/relation} between them, rather than about individual instances with possibly high-dimensional characteristics. 

We run a parallel experiment asking workers to guess  prices of $400$ flats, based on the full set of features our method is trained on. We assign one US-based worker per HIT, with approval rate greater than $97\%$ and over $500$ approved HITs, for a total of \$$12$ (including fees), $333\%$ higher than collecting ballpark group guesses. 

As we show below, the individual crowd estimations are not sufficient for training a good regression model. However, using guesses on groups in our ballpark methods achieves results comparable to a regression model based on \emph{true} apartment prices.

\xhdr{Aggregating crowd guesses of intervals}
In the above design, we asked workers to guess apartment prices and used percentiles of answers as bounds. While this worked well in practice, we seek an aggregation requiring less intervention by the practitioner.

As discussed in Section \ref{sec:Regression}, another approach is to  have the crowd directly \textbf{guess intervals}. Thus, in a separate experiment we ask people to guess lower and upper limits bracketing bag averages. To construct our constraints we then simply take the means of the upper and lower bounds, respectively. We formulate the task in simple language and encourage workers to take into account their uncertainty (``feel free to give a wide range if you are not sure'').

\xhdr{Results and robustness to outliers} In many real-world settings, data is often ``contaminated'' with observations that have outlier target values. In our data set, there is a small portion of apartments with very high prices in comparison to the rest (about $0.5$ percent of apartments are priced over \$$1000$).  These outlier apartments raise several points of interest. Unsurprisingly, people are not good at guessing the prices of outlier flats, rendering the labels particularly off-mark. More importantly, we find that while ridge regression suffers a considerable drop in accuracy due to these observations, our method is naturally robust since the crowd's guesses on \emph{groups} inherently disregard extreme, non-representative behaviors.

We compare our label-free method to ridge regression with a different number of true labels, and also to ridge using $400$ labels obtained from workers. To handle outliers, we use the Mean Absolute Error (MAE) metric rather than RMSE in $10$-fold CV (so that the unit of error is in dollars). We compare results running on the entire data, and removing all instances with prices over \$$1000$.

\begin{figure}[t]
	\includegraphics[width=0.85\linewidth]{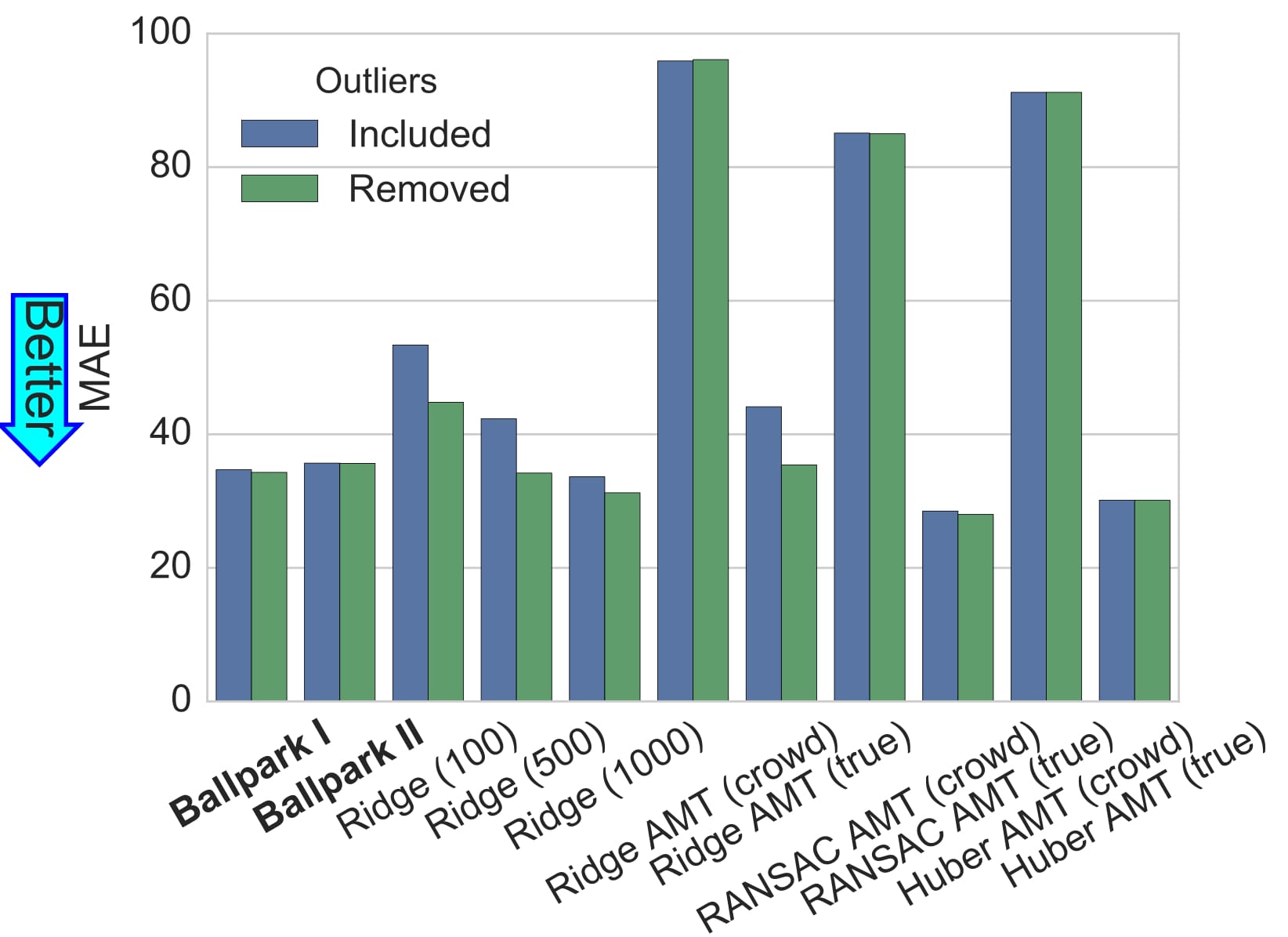}
	\caption{\small Airbnb results. Ballpark I denotes percentile aggregation of guesses, Ballpark II denotes eliciting interval predictions. Ridge regression is trained with an increasing number of labels. Ridge with $400$ \textit{individual} labels from crowd guesses performs poorly, as opposed to using constraints on groups in our Ballpark framework, or training ridge on the same $400$ instances with \textit{true} labels. Similar results hold for the RANSAC and Huber robust regression models. \remove{Both constraint aggregation approaches (Ballpark I and II) lead to nearly identical results, while approach is II is slightly more hands-off.} \label{fig:air_res}}
\vspace{-0.5cm}
\end{figure}

As seen in Figure \ref{fig:air_res}, our method achieves MAE results comparable to ridge regression with a large number of true labels. While outliers cause a big increase in error for ridge regression with $100$ labels, our method remains nearly unaffected. Our method is able to near results reached with robust regression models (RANSAC \cite{fischler1981random}, Huber regression \cite{huber2011robust}) trained on ground-truth labels.

To further test the effect of outliers and the possibility that the poor baseline ridge results are due to a few large errors, we train RANSAC and Huber regression with the $400$ crowd labels. These methods manage to only slightly reduce error, showing that while outliers have \textit{some} effect, the overall quality of crowd labels is the key source of error, which our bag-based method is able to avoid.  

Interestingly, both constraint aggregation approaches -- using percentiles (Ballpark I in in Figure \ref{fig:air_res}) and eliciting interval predictions directly (Ballpark II) -- lead to nearly identical results, the latter being more hands-off.  
We also see that ridge regression with $400$ crowd-acquired labels does rather poorly, although it does well for the exact same instances when their true labels are given.

\section{Related work}
\label{sec:RelatedWork}
\xhdr{Crowdsourcing for machine learning} There is a large body of work about the use of crowdsourcing for machine learning, primarily regarding label collection. In Section \ref{sec:CrowdRelated} we reviewed the most relevant work. The main focus in that field is acquiring \emph{discrete} labels for classification. Getting \emph{accurate} labels typically requires a lot of resources: multiple queries, worker reputation, and probabilistic models of worker patterns (e.g., biases). Our approach is different: First, we exploit the natural
human tendency for intuiting on \emph{groups} and the tendency
for \emph{comparisons}. Second, instead of focusing resources on aggregating crowd guesses accurately, we leverage a machine learning model based on rough intervals bracketing label averages, to accurately predict individual labels with \emph{few resources}. In addition, we handle \emph{regression} and continuous targets, which is notoriously hard for crowds and has not seen much work.

\xhdr{Related learning settings}
The field of \emph{Multiple Instance Learning} (MIL) is concerned with ``bags" of instances, where each bag has a label associated with it. This label is modeled as a function of latent instance-level labels, which can be seen as a form of weak supervision. MIL methods vary by the assumptions made on this function \cite{Cheplygina14bags,Foulds10MIL}. Most work in MIL focuses on making bag-level predictions rather than for individual instances. \remove{Recently, \cite{Kotzias15deep} used a convolutional neural network to predict labels for sentences given document-level labels.} 
%
In a related line of work, Learning from Label Proportions, individual labels are predicted based on known label proportions for bags \cite{quadrianto2009estimating,Ruping10svmclassifier,Felix13svm}.
In \cite{Ruping10svmclassifier}, each bag is represented with its mean, \remove{and class proportions are modeled based on this representative ``super-instance"}showing superior performance over \cite{quadrianto2009estimating}. In \cite{Felix13svm}, individual labels are explicitly modeled to counter problems with representing bags by their means (such as high variance).
These approaches all assume bag proportions are known, an assumption Ballpark Learning relaxes. 

To the best of our knowledge, the  subject of continuous labels and regression is not discussed in this literature, let alone demonstrated on data (simulated or real). In \cite{quadrianto2009estimating} the authors mention that their framework could apply, in theory, to continuous label spaces, but their methods assume a discrete label space to be able to reconstruct class probabilities efficiently.

\remove{Finally, the seminal work of \cite{xing03side} uses side-information for clustering in the form of pairwise constraints on cluster membership. Much work has since been done along these lines. We incorporate pairwise constraints in our methods, with pairs representing bags of instances, and partial ordering with respect to label averages.}
\remove{Finally, the seminal work of \cite{xing03side} on clustering uses pairwise constraints on cluster membership. We use pairwise constraints on bags of instances, with partial ordering based on label averages.}

\remove{ 
\xhdr{Robust Optimization}
Finally, robust optimization \cite{ben2009robust} research deals with uncertainty-affected optimization problems, by optimizing for the \emph{worst-case} value of parameters. Because of its worst-case design, robust optimization can do poorly when the constraints are not tight. Our method, on the other hand, is designed to handle rough estimates and loose constraints.}

\section{Conclusion and Future Work}

In this work we proposed a new method that can complement standard crowdsourcing practices when traditional labeling is difficult. People often have intuition about groups of instances and relations between them, while labeling individual instances is hard.
Our framework takes advantage of this phenomenon, leveraging a recent machine learning setting called Ballpark Learning based on weak, noisy constraints over \emph{groups} of instances. We extended Ballpark Learning to handle the useful case of continuous outputs, formulating a convex program with a simple solution. Across several real datasets,  we harness constraints from a crowd of non-experts and use them to train learning models. Our results rival supervised models that use many \emph{true} labels, at a much lower cost. 

In practice it may be unclear how to construct useful bags. Interesting future work is using crowdworkers to select and build bags \textit{themselves}, perhaps giving them a GUI to explore slices of the data. 
\remove{From a practical point of view, it may be unclear how to construct useful bags. Interesting future work is using crowdworkers to select and build bags \textit{themselves}, perhaps by giving them a visualization GUI to explore slices of the data.}

Deriving a deeper theoretical analysis of our models is also interesting. For example, understanding what makes bags ``useful'' in terms of the signal they provide, depending on factors such as their size and dispersion. The Ballpark approach can also potentially be combined with deep learning models (typically requiring many labels) as a form of weak supervision. We believe our lightweight methods pose an interesting alternative to current labeling practices, and could be particularly useful when data is high-dimensional and unintuitive for crowds, when privacy concerns prevent showing individual examples, and when resources are limited.

\xhdr{Acknowledgments} The authors thank the anonymous reviewers for their helpful feedback. Dafna Shahaf is a Harry \& Abe Sherman assistant professor. This work was supported by ISF grant 1764/15, Alon grant, and the HUJI Cyber Security Research Center in conjunction with the Israel National Cyber Bureau in the Prime Minister's Office.


\newpage

\bibliographystyle{ACM-Reference-Format}
\balance
\bibliography{sigproc} 
\remove{\begin{lemma}
For $\mathbf{w} \in \mathbb{R}^d$, and under some standard simplifying assumptions, Problem 
\ref{eq:prob_init_reg} can be cast in the general PAC learning model, and we obtain the basic sample complexity bound of 
\begin{equation}
\begin{aligned}
m_{\mathcal{H}} (\epsilon,\delta) \leq \frac{128d + 2\log(\frac{2}{\delta})}{\epsilon^2}, 
\end{aligned}
\end{equation}
where $\mathcal{H}$ is an hypothesis class $\mathbf{w}$ induces over instances $\mathbf{x}$ and bags $\mathcal{B}$, and $\epsilon,\delta$ follow the standard PAC notation (see \cite{shalev2014understanding}).  
\end{lemma}	

We begin by sketching how a slightly simplified version of our problem can be cast in the general PAC learning model. This will allow us to derive basic sample complexity bounds.

First, to simplify our exposition we set $C_L = 0$ as in our experiments. This means we do not have access to any labels. In addition, for ease of deriving basic sample complexity bounds, we drop the regularization term (we discuss this point shortly below). 

Now, we can immediately observe that every $\mathbf{w}$ induces a hypothesis over individual instances: $\mathbf{x} \mapsto \mathbf{w}^{T}\varphi(\mathbf{x})$. $\mathbf{w}$ also induces a hypothesis over bags:  $\mathcal{B} \mapsto  \frac{\sum_{i \in \mathcal{B}_{k}} \mathbf{w}^{T}\varphi(\mathbf{x_i})}{|\mathcal{B}|}$.

Next, we can incorporate our constraints into a loss function, as follows. Given a bag $\mathcal{B}$ and upper and lower bounds $u, l$ on $\frac{\sum_{i \in \mathcal{B}} \mathbf{w}^{T}\varphi(\mathbf{x})}{|\mathcal{B}|}$, we can obtain a $0$-$1$ loss function from constraints:
  $l(\mathbf{w}; (\mathcal{B} , l, u)) = 0$ if  $l \leq  \frac{\sum_{i \in \mathcal{B}} \mathbf{w}^{T}\varphi(\mathbf{x})}{|\mathcal{B}|} \leq u$, and otherwise the loss is $1$.
Finally, given a distribution $\mathcal{D}$ over triplets $(\mathcal{B}, l, u)$, we define the population risk by $L_{\mathcal{D}}(\mathbf{w}) =\mathbb{E}_{(\mathcal{B}, l, u) \sim \mathcal{D}} \lbrack l(\mathbf{w}; (\mathcal{B} , l, u)) \rbrack$. We have now cast the problem in the general PAC learning model.

Under the PAC framework, we can get a simple sample complexity result. In the literature of learning theory, there are dimension-based bounds and norm-based bounds. For simplicity, we derive a dimension-based bound using the ``discretization trick'' \cite{shalev2014understanding}. Denote our hypothesis class (as specified above) with $\mathcal{H}$. $\mathcal{H}$ is parameterized by $\mathbf{w} \in \mathbb{R}^d$. This is an infinite hypothesis class. However, practically speaking, we can represent each of these $d$ numbers in a computer with a (say) $64$-bit floating point number. Thus the size of our hypothesis class is at most $2^{64d}$. Using standard results (corollary $4.6$ in \cite{shalev2014understanding}), we get that the sample complexity $m_{\mathcal{H}} (\epsilon,\delta)$ for our problem is bounded by $\frac{128d + 2\log(\frac{2}{\delta})}{\epsilon^2}$, where $\epsilon,\delta$ follow the standard PAC notation (see \cite{shalev2014understanding}).  

Using standard (but tedious) covering number arguments, one can derive a norm-based bound. In practice, we found out that regularizing over $\mathbf{w}$ (corresponding to norm-based bounds) leads to better empirical results.}

\end{document}